\documentclass[preprint]{article}

\usepackage{neurips_2026}

\usepackage[utf8]{inputenc} 
\usepackage[T1]{fontenc}    
\usepackage{hyperref}       
\usepackage{url}            
\usepackage{booktabs}       
\usepackage{amsfonts}       
\usepackage{nicefrac}       
\usepackage{microtype}      
\usepackage{xcolor}         
\usepackage{graphicx}
\usepackage{amsmath}
\usepackage{float}

\usepackage{enumitem}

\usepackage{algorithm}
\usepackage{algpseudocode}

\usepackage{comment}

\newcommand{\method}{\textsc{Retokenization Sampling}}
\newcommand{\passretok}{\mathrm{pass@retok}}
\newcommand{\passk}{\mathrm{pass@k}}
\newcommand{\pass}{\mathrm{pass}}
\newcommand{\passtypos}{\mathrm{pass@typo}}

\title{Emergent retokenization symmetry in large language models: phenomenology and applications}

\author{%
  Kanishk ~Jain$^{*}$\quad
   Matthew Day \quad 
  Tankut Can$^{*}$ \\
  Department of Physics, Emory University, Atlanta, GA 30322 \\
  $^{*}$ Correspondence: \texttt{\{kanishk.jain,tcan\}@emory.edu} \\
}

\begin{document}

\maketitle

\begin{abstract}

Tokenization introduces representational redundancy: under a fixed token vocabulary, every byte string admits many valid token encodings, or segmentations, that decode to the same surface string. However, given a prompt, most language model tokenizers break this representational symmetry by returning a canonical segmentation. Training only on canonical segmentations should influence model behavior at inference, and there is little reason to expect the model to respect segmentation symmetry on downstream tasks. Nevertheless, we find that this symmetry partially emerges during training. Here, we probe this emergent symmetry through a series of experiments testing token compositional understanding, representation diversity, and task focused benchmark performance. We primarily use \textbf{retokenization} -- replacing a prompt's canonical tokenization with an alternative valid segmentation while preserving its bytes exactly. Relative to other prompt perturbations, retokenization is unusually clean because it isolates segmentation effects without changing syntax, semantics or surface form. We use retokenization to study sensitivity and robustness to semantically identical input representations across pretraining and post-training. Moreover, this partial retokenization symmetry suggests a distinct inference-time sampling axis. While temperature sampling generates diverse outputs from the model using its next-token probability distribution, retokenization generates diversity from the model's internal computations through semantically equivalent input representations. We find that while this retokenization sampling strategy can hurt performance on easy problems, it can also recover solutions that conventional sampling does not find. Overall, our work presents retokenization as a simple yet powerful probe of large language models, shedding light on compositional understanding and prompt sensitivity, and offering a novel sampling strategy.

\end{abstract}

\section{Introduction}

Tokenization is usually discussed as a source of brittleness in large language models (LLMs): grouping multiple characters into a single token obscures character structure and can produce failures on spelling-sensitive or character-level tasks \citep{singh_tokenization_2024, edman_cute_2024, cosma_strawberry_2025,chai_tokenization_2024}. The naive picture often called upon to explain these phenomena is that learning from token sequences does not necessarily produce a full understanding of the constituent structure of tokens. But this picture is incomplete, as recent work shows that LLMs can often interpret alternative segmentations of the same string and may internally recover whole-word representations from subword sequences \citep{zheng_broken_2026, kaplan_tokens_2025}, suggesting an emergent form of compositional understanding across token boundaries. Nevertheless, residual differences remain between different segmentations, affecting computation and performance \cite{geh_adversarial_2025}.

With a fixed vocabulary, the model's \textbf{canonical tokenizer} returns only one valid encoding of a given byte string using either dynamic programming or a greedy decoding algorithm. However, given a \textit{fixed vocabulary of tokens}, many alternate \textbf{segmentations} can be created, each represented by a different sequence of tokens that decode to the same surface string (Figure \ref{fig:fig2}B). We call the alternative non-canonical segmentations {\bf retokenizations}. For instance, the token \fbox{\texttt{probable}} can be represented as a two-token sequence \fbox{\texttt{prob}} \fbox{\texttt{able}}, if the tokens \fbox{\texttt{prob}} and \fbox{\texttt{able}} are present in the model's existing vocabulary. Using the canonical tokenizer during training constrains an LLM's ability to learn such composition of tokens. This blindness to subtoken structure has been a source of embarrassment in the past for frontier models which could perform complicated mathematical and coding tasks, but seemingly could not count the letters in a word \citep{cosma_strawberry_2025, chai_tokenization_2024}. 

If training on canonical segmentations produced no invariance at all, non-canonical segmentations would simply corrupt the prompt. If invariance were exact, all equivalent segmentations would induce the same computation and retokenization would be behaviorally trivial. The empirical regime of interest is the intermediate one: trained models may exhibit approximate {\bf segmentation symmetry}, meaning they still understand non-canonical segmentations \citep{kaushal2022tokensknowcharactersknow, itzhak_models_2022, edman_cute_2024}, together with residual {\bf segmentation sensitivity}, meaning alternative segmentations of the same prompt produce different internal states and outputs \citep{zheng_broken_2026, cosma_strawberry_2025, edman_cute_2024}.

To probe this symmetry--sensitivity tradeoff, we use retokenization as a controlled inference-time intervention. Given a fixed vocabulary and byte-level prompt $s$, let $E^0(s)$ be the canonical token encoding and let $E^\mu(s)$ be any other valid segmentation of $s$ under the same vocabulary. Retokenization never changes the prompt bytes, introduces no new tokens, and does not define a new task. It changes only how the same string is decomposed into \textit{existing} token  vocabulary items. We use this perturbation in two ways throughout the paper. First, we use it mechanistically, to study hidden-state changes under semantically identical inputs. Second, we use it behaviorally through \method{}, which samples segmentations $\{E^\mu(s)\}_{\mu=1}^k$, decodes greedily from each one, and summarizes the resulting success curve with $\passretok(k)$, the analogue of $\passk$ in which diversity comes from input representation rather than output randomness.

This setup is intentionally narrower than generic prompt perturbation. Paraphrasing, synonym substitution, reframing, masking and reformatting change the surface string and therefore confound semantics with robustness to wording \citep{liu_pre-train_2021, qiang_prompt_2024, bsharat_prompting_2025, dang_data_2025}. Retokenization, by contrast, changes only the token-level representation of a fixed byte string. This makes it a particularly clean probe of whether tokenized models process equivalent inputs in a segmentation-invariant way, and where that invariance breaks down. Across HumanEval \citep{chen_evaluating_2021},  GSM8K \cite{cobbe_training_2021}, GSM8K Python \citep{chowdhery_palm_2022}, and MMLU \cite{hendrycks_measuring_2021} benchmarks, we find that equivalent segmentations usually preserve task identity and generate nontrivial diversity in outputs. Moreover, across OLMo-2 pretraining and post-training, we find that byte-level understanding, necessary for segmentation symmetry, is gradually learned rather than present trivially.

Our approach is distinct from prior methods that use stochastic tokenization during training, where subword regularization encourages robustness by exposing multiple segmentations during training \citep{kudo_subword_2018, provilkov_bpe-dropout_2020}. We mainly focus on understanding inference-time behavior in models that were not explicitly trained for segmentation invariance. In that setting, any useful diversity produced by equivalent segmentations is an emergent property of the trained model rather than a built-in invariance constraint. 

This approach also connects tokenization to recent views of reasoning as trajectory selection under finite compute \citep{snell_scaling_2024, merrill_expressive_2024, pfau_lets_2024}. Transformers compute token by token, so changing segmentation changes the granularity and ordering of computation within the model even when the prompt string is unchanged. In that sense, retokenization is a controlled method of altering the model's internal processing without changing the prompt itself. The intriguing mechanism is therefore not just whether alternative segmentations help or hurt benchmark performance, but what they reveal about how invariance and sensitivity coexist inside tokenized models. 

This question is timely because tokenization research increasingly treats tokenizers as more than compression devices \citep{bostrom_byte_2020, zouhar_tokenization_2023, schmidt_tokenization_2024, ali_tokenizer_2024, haslett_tokenization_2025}, while byte-level and tokenizer-free models aim to remove the dependency altogether \citep{xue_byt5_2022, yu_megabyte_2023, wang_mambabyte_2024, pagnoni_byte_2024}. Our work addresses the complementary regime: as long as deployed LLMs remain tokenized, the multiplicity of valid segmentations is itself a mechanistically meaningful degree of freedom. In the very least, it cannot be ignored. 

The rest of the paper develops this argument in three steps. We first ask whether segmentation-level compositional understanding emerges during training, using hidden state analyses. We then measure the behavioral consequences of residual segmentation sensitivity across four benchmarks and multiple training stages. Finally, we compare the diversity induced by retokenization to more conventional sources of output variation to compare segmentation-induced diversity to standard baselines. 

\subsection{Our Contributions}

We make the following contributions:

\begin{enumerate}[leftmargin=*, label=\arabic*.]

\item \textbf{Retokenization as a controlled probe of segmentation invariance.} We formalize retokenization as a semantics-preserving intervention that changes only the token-level representation of a fixed byte string. This gives a clean way to probe how tokenized models respond to equivalent inputs without changing wording, syntax, or task definition.

\item \textbf{Internal representations show emergent segmentation symmetry and residual sensitivity.} Using hidden-state analyses, we show that segmentation-induced variation is suppressed in early and middle layers but is accentuated in the final layer.

\item \textbf{A concrete evaluation protocol.} We formalize the symmetry--sensitivity tradeoff with $\passretok(k)$, the direct analogue of pass@k in which diversity comes from input representation rather than output randomness. This provides a concrete way to study how segmentation invariance and sensitivity coexist in tokenized language models.

\item \textbf{Task-dependent gains and robustness trends.} Across GSM8K, GSM8K Python, and HumanEval, we show that $\passretok(k)$ rises with $k$ and recovers instances missed by canonical decoding. We also show that across post-training stages in OLMo-2 and other model families, stronger models also tend to remain stronger under retokenization. In other words, retokenization symmetry tends to track model performance.

\item \textbf{Retokenization recovers structurally diverse correct programs.} On HumanEval, AST-based syntactic-diversity measurements show that retokenization does not merely reproduce a single canonical solution template. Correct retokenized generations span a substantial range of program structures, and some tasks are solved only by \method{}.

\end{enumerate}

\section{Related Work}

Subword tokenization schemes such as BPE, WordPiece, and SentencePiece were introduced to balance open-vocabulary coverage with computational efficiency \citep{sennrich_neural_2016, kudo_sentencepiece_2018, schuster_japanese_2012, radford_language_2019}. But tokenization is not just a compression step applied before modeling begins, as the choice of vocabulary and token boundaries changes how text is presented to the model, which features are localized within single tokens versus split across multiple tokens, and how much sequence length is required to express the same input. Prior work shows that these design choices affect pretraining efficiency, downstream performance, and the kinds of linguistic structure models capture reliably \citep{bostrom_byte_2020, ali_tokenizer_2024, zouhar_tokenization_2023, schmidt_tokenization_2024}. Tokenization is therefore an important part of building language models and cannot be treated as a detail that is irrelevant to model behavior.

In modern LLMs, prior works have found partial but incomplete character and sub-token level awareness \citep{kaushal2022tokensknowcharactersknow, itzhak_models_2022, edman_cute_2024}. Other studies show evidence of better than expected handling of non-canonical segmentations \citep{zheng_broken_2026}, alongside systematic failures in character level and arithmetic tasks \citep{singh_tokenization_2024, cosma_strawberry_2025}. Recent mechanistic work studies character-level tokenization and argue that robustness arises by early-layer in-group attention that reconstructs canonical lexical units from fragmented inputs \citet{yang_word_2026}. Most closely related, \citet{geh_adversarial_2025} show that non-canonical segmentations can preserve semantics strongly enough to bypass safety filters. Our use of retokenization is narrower and constructive: we treat it as a semantics-preserving probe of both robustness and inference-time diversity. 

Across input perturbation methods such as paraphrases, formatting choices such as delimiters and whitespace, verbalizers, and typos, LLMs have been shown to be particularly sensitive to small changes \citep{liu_pre-train_2021, qiang_prompt_2024, bsharat_prompting_2025, sclar_quantifying_2024, salinas_butterfly_2024, errica_what_2025, zhuo_prosa_2024, gan_reasoning_2024, zhu_promptrobust_2024, wahle_paraphrase_2024}. To mitigate tokenization induced input sensitivity, subword regularization methods such as BPE-dropout expose models to multiple segmentations during training \citep{kudo_subword_2018, provilkov_bpe-dropout_2020}, while byte-level and tokenizer-free models avoid subword segmentation altogether \citep{xue_byt5_2022, yu_megabyte_2023, wang_mambabyte_2024, pagnoni_byte_2024}. Parallel work modifies tokenization for efficiency or adaptation \citep{feher_retrofitting_2025, geng_zip2zip_2025, pilana_liyanage_adaptbpe_2026, kwiatkowski_fusion_2023}, while multilingual and morphology-aware work has shown that tokenization can have fairness and language dependent consequences \citep{rust_how_2021, hofmann_superbizarre_2021, petrov_language_2023, limisiewicz_myte_2024}.

Despite this broad evidence that surface form matters, most studies measure sensitivity only behaviorally, through accuracy, attack success rate, or efficiency. Much less is known about whether meaning-preserving perturbations also change the model’s internal representations. A few exceptions use representation alignment as a robustness objective \citep{agrawal_enhancing_2025}, but they do not characterize how segmentation itself changes representations. \method{} fills this gap by perturbing only the segmentation while preserving the byte string exactly. Unlike typos, paraphrases, or formatting changes, this gives a controlled way to study how tokenized models process equivalent inputs, and our analysis is, to our knowledge, the first to characterize segmentation-induced prompt perturbations at the representational level.

\section{Results}

\subsection{Emergence of Segmentation Symmetry in Representation Space}\label{sec:segsym}

\begin{figure}[h]
  \centering
  \includegraphics[width=\textwidth]{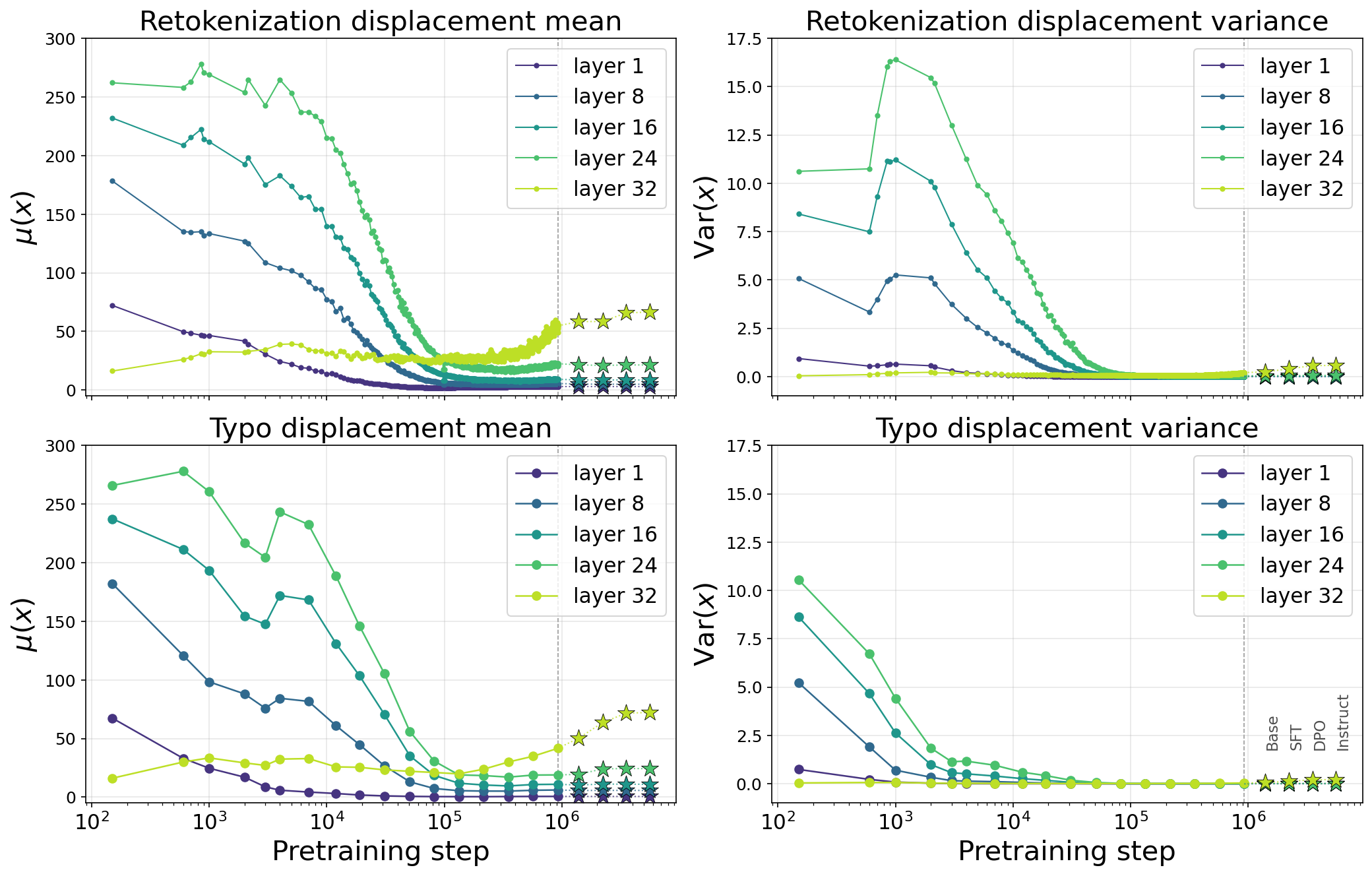}
  \caption{\textbf{Statistics of Latent Representations under retokenization and typo perturbations.} Retokenization of a prompt displaces latent reps relative to the canonically tokenized prompt. We show the mean \ref{eq:mu} ({\bf Top Left}) and variance \ref{eq:var} ({\bf Top Right}) of this displacement for OLMo-2-7B during pretraining, for a selection of layers. For comparison, we show behavior for typo pertubations, plotting mean ({\bf Bottom Left}) and variance ({\bf Bottom Right}). For both prompt perturbations, the displacement mean converges in all layers except the final layer 32, where it grows. Note that for OLMo-2-7b, layer 32 is the final layer before the unembedding layer, and is related to layer 31 only by an RMSnorm operation. To generate retokenizations and typos, we use $p_{retok} = p_{typo} = 0.5$. }
  \label{fig:latents}
\end{figure}

How do LLMs process non-canonical segmentations? To answer this, we study internal (latent) representations layer-wise, and throughout pretraining. More precisely, we explore representations of retokenization in intermediate layers of OLMo-2-7B during pretraining. We provide the model with the input prompt from a selection of HumanEval prompts, and calculate the hidden state of the final token in the prompt, at each layer of the model. Denote this final-token hidden state at layer $\ell$ induced by prompt $E^{\mu}(s)$ by ${\bf h}_{\ell}^{\mu} \in \mathbb{R}^{D}$. We write the components of this vector as $h_{\ell,i}^{\mu}$ for $i =1, ..., D$. We calculate the displacement from the canonical representation, 
\begin{align}
    {\bf x}_{\ell}^{\mu} =  {\bf h}_{\ell}^{\mu} - {\bf h}_{\ell}^{0}, 
\end{align}
where the components are denoted $x_{\ell,i}^{\mu}$. In Figure \ref{fig:latents} we report the norm of the average displacement $\mu_{\ell}(x)$, as well as the variance of displacement ${\rm Var}_{\ell}(x)$, for different layers: 

\begin{align}
   \mu_{\ell}(x) &\equiv ||\langle  {\bf x}_{\ell}\rangle||, \quad \langle  x_{\ell,i}\rangle \equiv  \frac{1}{S}\sum_{\mu=1}^{S} x^{\mu}_{\ell,i},\label{eq:mu}\\
   {\rm Var}_{\ell}(x) &\equiv  \frac{1}{(D-1)S} \sum_{i = 1}^{D}  \sum_{\mu = 1}^{S} \left( x_{\ell,i}^{\mu} - \langle x_{\ell,i}\rangle \right)^{2} \label{eq:var} .
\end{align}

As a baseline prompt-perturbation comparison, we also look at displacement under prompts to which typos are added (for details see \ref{app:typo}). In this setting, the canonical prompt is the same as before $E^{0}(s)$, while the perturbed prompt has typos injected and is canonically tokenized. The latent representations are computed in the same way.

To summarize, under retokenizations, a model's internal representations pick up a bias and variance. Both of these summary statistics significantly diminish during training, but do not completely collapse. Therefore, while the internal representations cluster non-canonical segmentations, they still encode differences. We explore this residual sensitivity to segmentation in reasoning taks below. 

\subsection{Consequences of Residual Segmentation Sensitivity on Reasoning Tasks}

The results from previous section suggest that retokenized prompts are assigned different representations inside the model. But do these differences survive in reasoning traces? In other words, do the initial differences in representations lead to different outputs? To answer this, we use \method{} across benchmarks in three distinct domains: mathematical reasoning (GSM8K), code generation (HumanEval) and code-mediated mathematical reasoning (GSM8K Python). We find that sampling alternative segmentations of a fixed prompt produces useful but task-dependent diversity. Concretely, \method{} takes a fixed prompt string, samples $k$ valid non-canonical segmentations of that same string under the model's existing vocabulary, runs the model once on each segmented prompt, and decodes greedily from each run. 

\begin{figure}[h]
  \centering
      \includegraphics[width = \textwidth]{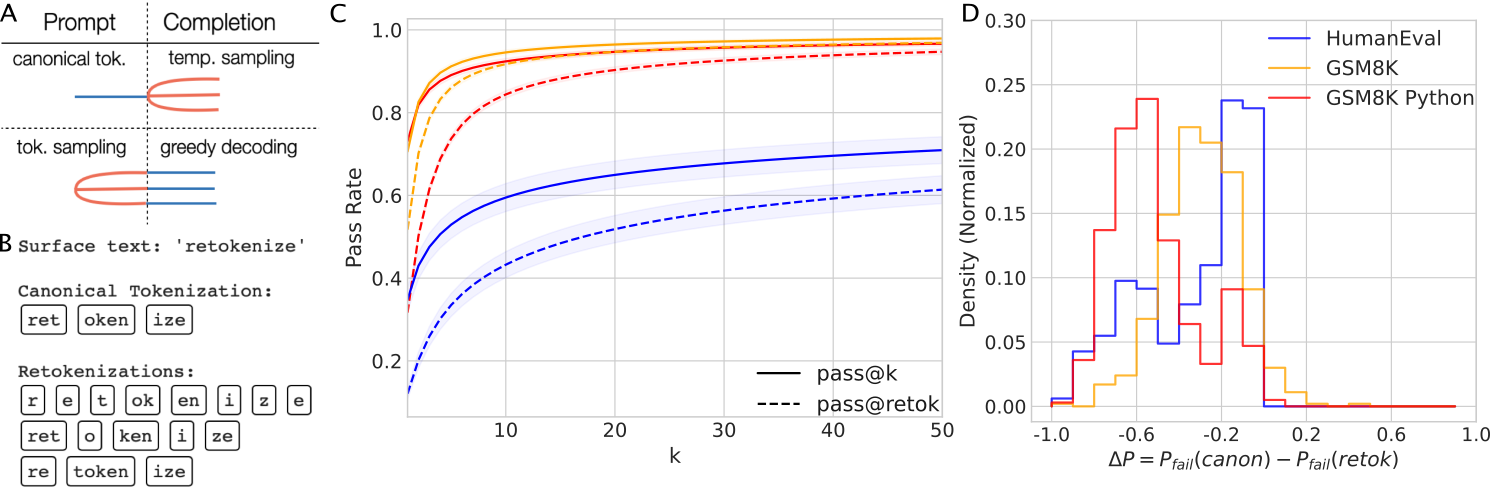}
  \caption{\textbf{Retokenization sampling as an input-side search axis.} \textbf{A)} Conventional sampling fixes the canonical prompt encoding and samples from the output distribution (at temperature = 1), whereas retokenization sampling varies the prompt segmentation and decodes greedily. \textbf{B)} A single surface string admits many valid tokenizations under a fixed vocabulary; the canonical tokenizer chooses only one of them. \textbf{C)} Across HumanEval, GSM8K, and GSM8K Python, $\passretok$ exhibits the same qualitative behavior as $\passk$, showing that different segmentations expose distinct useful trajectories. The shaded area shows standard errors (see Appendix~\ref{app:passk_errorbars}). \textbf{D)} Histogram of per-problem failure-probability differences, $\Delta P = P(\mathrm{fail} \mid \mathrm{canon}) - P(\mathrm{fail} \mid \mathrm{retok})$. Mass above zero indicates problems helped by retokenization. All experiments presented here are for OLMo-2 7B Instruct model.}
  \label{fig:fig2}
\end{figure}

We quantify the output behavior with $\passretok(k)$, a metric designed to parallel the standard $\passk$ evaluation from \citet{chen_evaluating_2021}: for a fixed benchmark prompt, we ask how often at least one segmentation out of $k$ gives a correct solution. Because decoding is deterministic, variation in $\passretok(k)$ reflects diversity induced by the input representation rather than by output sampling. Figure \ref{fig:fig2}C shows that these retokenization-sampling curves have the same qualitative rising-with-diminishing-returns shape as $\passk$ across all four benchmarks, with the largest gap to canonical $\passk$ on HumanEval and smaller gaps on GSM8K and GSM8K Python. GSM8K Python falls naturally between GSM8K and HumanEval: the question remains natural language, but the answer space is programmatic. Figure \ref{fig:fig2}D shows that these aggregate gains are not uniform: the distribution of $\Delta P = P(\mathrm{fail}\mid \mathrm{canon}) - P(\mathrm{fail}\mid \mathrm{retok})$ places noticeable mass above zero, indicating that retokenization reduces failure probability on a subset of instances rather than only changing the average curve.

\begin{figure}[h]
  \centering
  \includegraphics[width=\textwidth]{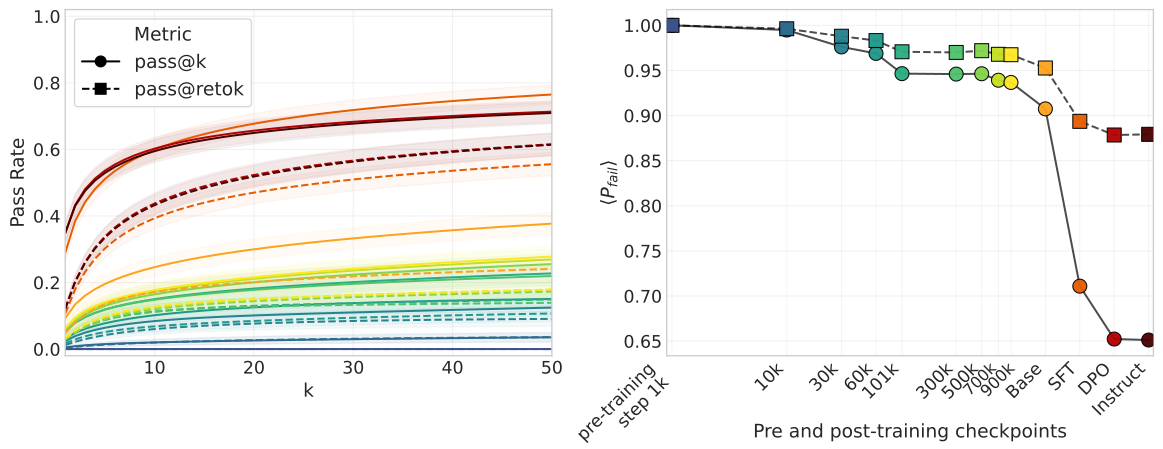}
  \caption{\textbf{Training improves canonical and retokenized sampling together.} \textbf{Left:} Solid curves show $\passk$ and dashed curves show $\passretok$ for successive OLMo-2 7B pre-training checkpoints and post-training stages on HumanEval. Each training step shifts both curves upward, with especially large gains from base to SFT and further gains through DPO. There appear marginal gains from DPO to Instruct. \textbf{Right:} the corresponding mean per-problem failure probability $\langle P(\mathrm{fail})\rangle$ decreases across training for both sampling procedures, but canonical sampling remains consistently lower than retokenized sampling. Segmentation robustness therefore tracks overall task competence rather than emerging as a separate skill.}
  \label{fig:training_humaneval}
\end{figure}

{\bf Change in segmentation sensitivity and robustness across training} We extend our experiments to track the same OLMo-2 7B model family across pre-training and post-training by obtaining $\passk$ and $\passretok(k)$ for the HumanEval benchmark. These models belong to a sequence of training stages which include, in the order in which they are implemented: base pre-training, supervised fine-tuning (SFT), direct policy optimization (DPO), and reinforcement learning with verified rewards (Instruct). Figure \ref{fig:training_humaneval}(left) shows a gradual increase in both ${\rm pass@k}$ and ${\rm pass@retok}$ during pre-training, and a sharp increase in both metrics with post-training. The corresponding mean failure probabilities shown in Figure \ref{fig:training_humaneval}(right) show a synchronized two-step decline for both sampling methods, the first occurring in early pre-training (before step $100000$) and a prominent decline appearing at the post-training SFT stage, consistent with the $\passk$ and $\passretok(k)$ trends. While the evidence and mechanism of improvement in ${\rm pass@k}$ scores from post-training has been shown before \citep{chen_evaluating_2021, cobbe_training_2021, ouyang_training_2022}, the reason behind a proportionate increase in ${\rm pass@retok}$ scores remains unclear. These results indicate a dominant role of post-training in the emergence of segmentation invariance in language models. We also observe the same qualitative pattern across model families: stronger models generally remain stronger under retokenization, with family-specific variation in the gap to canonical sampling (Appendix~\ref{app:model_families}).

\subsection{Diversity in generations from retokenization and temperature sampling}

The previous section suggests that retokenization can be viewed as a source of noise, allowing for a potentially new sampling axis. However, is there any difference between retokenization and temperature sampling in the space of sampled outputs? To probe this question, we measure syntactic diversity directly in the space of generated programs from HumanEval dataset using OLMo2-7B-Instruct model.

We adapt a previously introduced syntactic diversity metric \citep{shypula_evaluating_2026} by parsing each program generated by the model into a Python abstract syntax tree (AST) and extracting a set of bottom-up subtrees of fixed height ($h=4$). We then compare two programs by pooling their extracted subtrees and measuring how many of those pooled subtrees are unique. If $S_1$ and $S_2$ denote the multisets of subtrees from two programs, we define the distance between the subtree multisets as
\begin{equation}
d_{AST}(S_1, S_2) = 1- \frac{2\left|S_1 \cap S_2\right|}{|S_1| + |S_2|}
\end{equation}

where the numerator counts distinct subtree patterns across both generations, and the denominator counts the total number of subtree instances contributed by the two generations. This is just a Jaccard distance rescaled to give values in the unit interval. The highest possible distance is unity, which is obtained when $S_{1}$ and $S_{2}$ have no elements in common. The distance is zero when the sets are identical. Identifiers, argument names, and literal values are canonicalized before subtree extraction, so the metric emphasizes structural rather than lexical differences.

\begin{figure}[h]
  \centering
  \includegraphics[width=\textwidth]{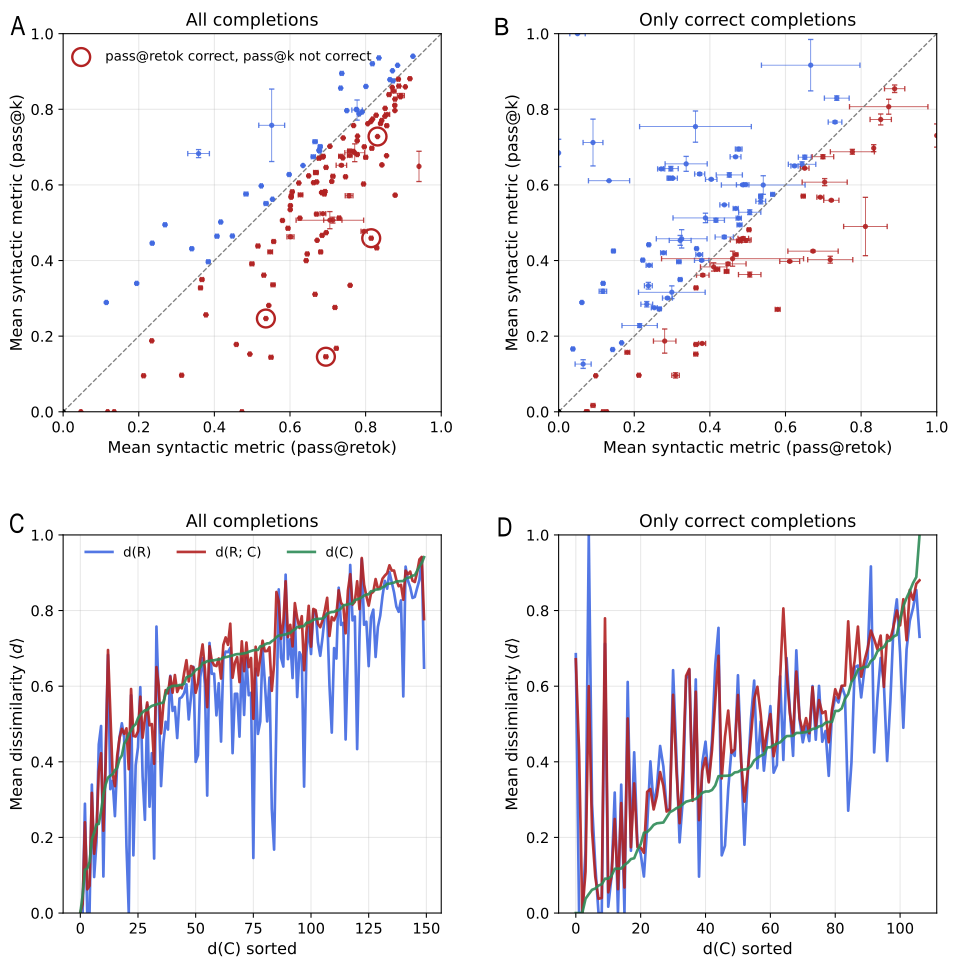}
  \caption{\textbf{Retokenization recovers structurally diverse correct programs (A,B).} Each point is a HumanEval task, comparing mean AST-based syntactic diversity under retokenization sampling (x-axis) and canonical temperature sampling (y-axis). Higher values reflect higher diversity (error bars reflect standard error). \textbf{A) }all completions. \textbf{B) }only correct completions. A strong correlation in diversity within correct completions shows that retokenization does not merely reproduce one correct solution; it explores a comparable range of program structures. Red circles highlight tasks only solved by \method{}.\textbf{Within-family and cross-family diversity are of similar scale (C,D).} Tasks are sorted by within-canonical diversity $d(C)$. Blue shows within-retokenization diversity $d(R)$, red shows cross-family diversity $d(C,R)$, and green shows within-canonical diversity $d(C)$. \textbf{C) }all completions. \textbf{D) }only correct completions. Retokenized completions typically remain near the canonical diversity scale rather than collapsing to a single template or diverging into a disjoint region of program space.}
  \label{fig:diversity}
\end{figure}

Figure~\ref{fig:diversity}A,B shows mean within-family syntactic diversity under temperature sampling and retokenization. The plotted mean for each task is the average of this pairwise score over all within-family completion pairs, and the error bars are standard errors over those pairwise values. When all completions are included, most tasks lie below the diagonal, indicating that per-task diversity is higher under retokenization sampling. Restricting to only correct completions per task shows a strong correlation, as these generations cover much of the same diversity range as correct temperature-sampled programs. The main takeaway is therefore not that retokenization matches temperature sampling exactly, but that it does not collapse onto a single canonical template, despite being greedily decoded. Instead, correct generations by retokenization often explore a substantial range of structurally distinct correct programs.

{\bf Comparing diversity in temperature sampling and retokenization sampling} While both temperature sampling and \method{} generate non-trivial amounts of syntactic diversity, we don't know if they are exploring the same solution families. A stronger comparison is therefore to look at the geometry of the two completion sets directly: for a fixed task, are retokenized programs mostly variations of the canonical temperature-sampled ones, or do they occupy neighboring but distinct regions of program space? We therefore add a \emph{cross-family} quantity: for each task, we compute $d_{\text{AST}}$ for every pair consisting of one temperature-sampled completion and one retokenization-sampled completion, and average over those cross-family pairs. Let $C$ denote the canonical temperature-sampled completions for a task and $R$ the retokenization-sampled completions for the same task, and write $d_{ij}$ for the pairwise $d_{\text{AST}}$ between completions $i$ and $j$. We then compare 

\begin{align}
d(C,R)&  \equiv  \langle d_{ij} \rangle_{i \in C, \,\,j \in R} \equiv  \frac{1}{|C| |R|}\sum_{i \in C,\,\, j \in R} d_{ij},\\
d(C) & \equiv  \langle d_{ij} \rangle_{i<j \in C},
\qquad
d(R)\equiv \langle d_{ij} \rangle_{i<j \in R}.
\end{align}

In Figure~\ref{fig:diversity}C,D, sorting tasks by $d(C)$ reveals that retokenized completions track the same coarse diversity ordering as canonical completions. For all completions, $d(R)$ and $d(C,R)$ usually remain close to $d(C)$, which means the retokenized solutions neither collapse onto a single canonical program family nor jump to a completely disjoint region of program space. On the subset of correct completions the curves become noisier, as expected from fewer samples, but the same qualitative conclusion survives. Cross-family diversity is of the same order as within-family diversity, so they don't separate into obviously disjoint regions of the measured program space. 

\subsection{\method{} comparison with typo sampling}

Typographical perturbations provide a useful control to our method because they also perturb input to the model, but unlike retokenization they do not preserve the exact byte string. A typo can perturb input representation by changing token boundaries, changing lexical cues, or by genuinely changing meaning. Retokenization only isolates the first mechanism, thus, typo sampling is a stronger but less controlled perturbation method: if it helps, it is harder to interpret why. We therefore treat it as a baseline for input-side diversity here rather than as a direct substitute for retokenization.
    
\begin{figure}[h]
  \centering
  \includegraphics[width=\textwidth]{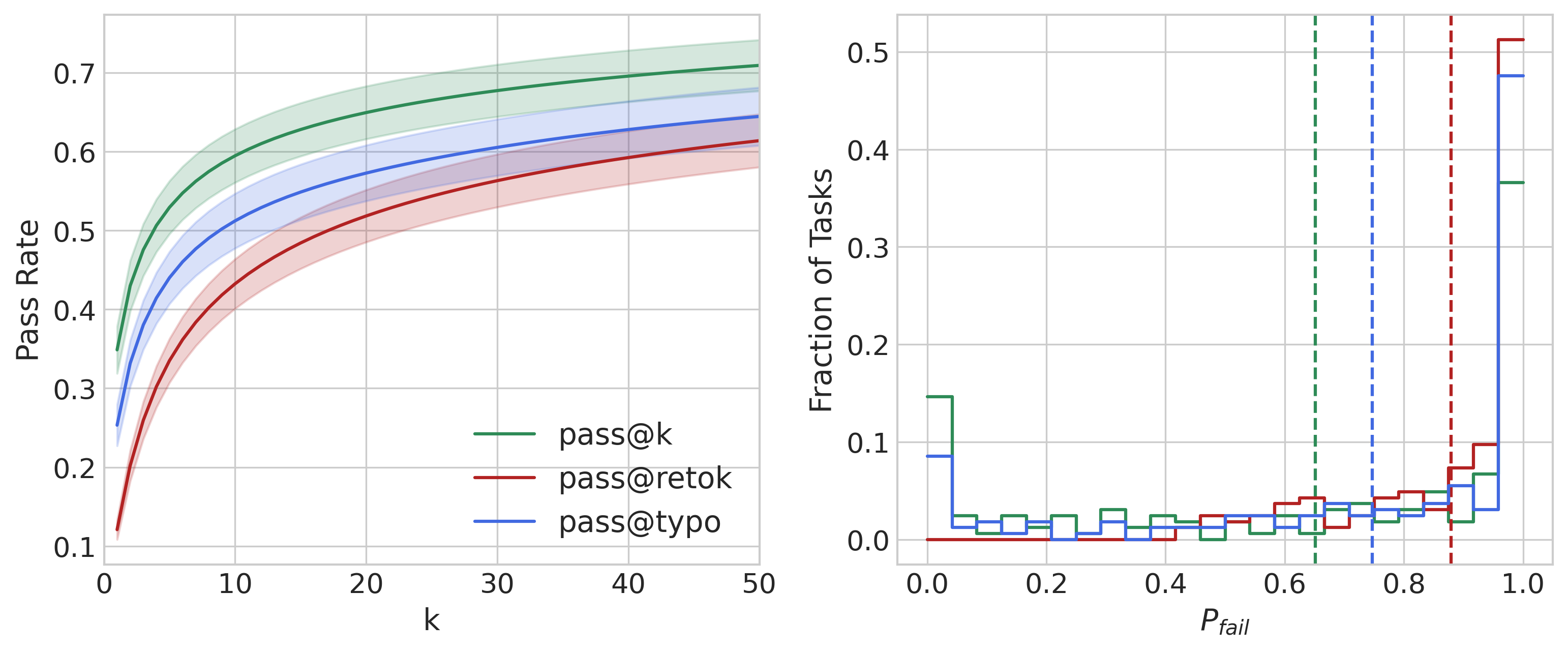}
  \caption{\textbf{Retokenization versus typo sampling on HumanEval.} \textbf{Left:} pass curves for temperature sampling (green), typo sampling (blue), and retokenization sampling (red) on HumanEval. Typo sampling consistently outperforms retokenization sampling but remains below conventional temperature sampling across the full range of $k$. \textbf{Right:} distributions over tasks of the empirical failure probability $P(\mathrm{failure})$ for the same three methods, with dashed vertical lines marking the corresponding means. Typo perturbations shift mass away from certain failure relative to retokenization, but still leave substantially more hard tasks than canonical output sampling. This ordering suggests that visible surface perturbations provide a stronger source of input-side diversity than segmentation alone, though they are also less controlled because they change the prompt text itself.}
  \label{fig:passtypo}
\end{figure}

Figure \ref{fig:passtypo} makes this control comparison on HumanEval dataset using OLMo-2 7B Instruct model. In the left panel, typo sampling (see Appendix~\ref{app:typo} for implementation details) consistently sits between canonical temperature sampling and retokenization sampling across the full range of $k$. The right panel shows the same ordering in a complementary way through the per-task failure-probability distributions: canonical sampling concentrates more mass on easy tasks with low $P_{fail}$, retokenization leaves the largest concentration near almost-certain failure, and typo sampling shifts some of that mass toward lower failure probabilities without matching canonical performance. One interpretation is that visible perturbations buy additional exploration power precisely because they move more than segmentation alone: they can alter lexical cues and local semantics, not just token boundaries. If typo sampling were helped only by re-segmentation effects, its distribution should be much closer to retokenization. Instead, the persistent gap suggests that a substantial part of typo-induced diversity comes from changing the prompt text itself. This strengthens the case for retokenization as the cleaner scientific probe. It is weaker as a search procedure, but when it changes outcomes we can attribute those changes specifically to segmentation rather than to new wording or corrupted semantics.

\subsection{Compute comparison across different sampling techniques}

\begin{figure}[h]
  \centering
  \includegraphics[width=\textwidth]{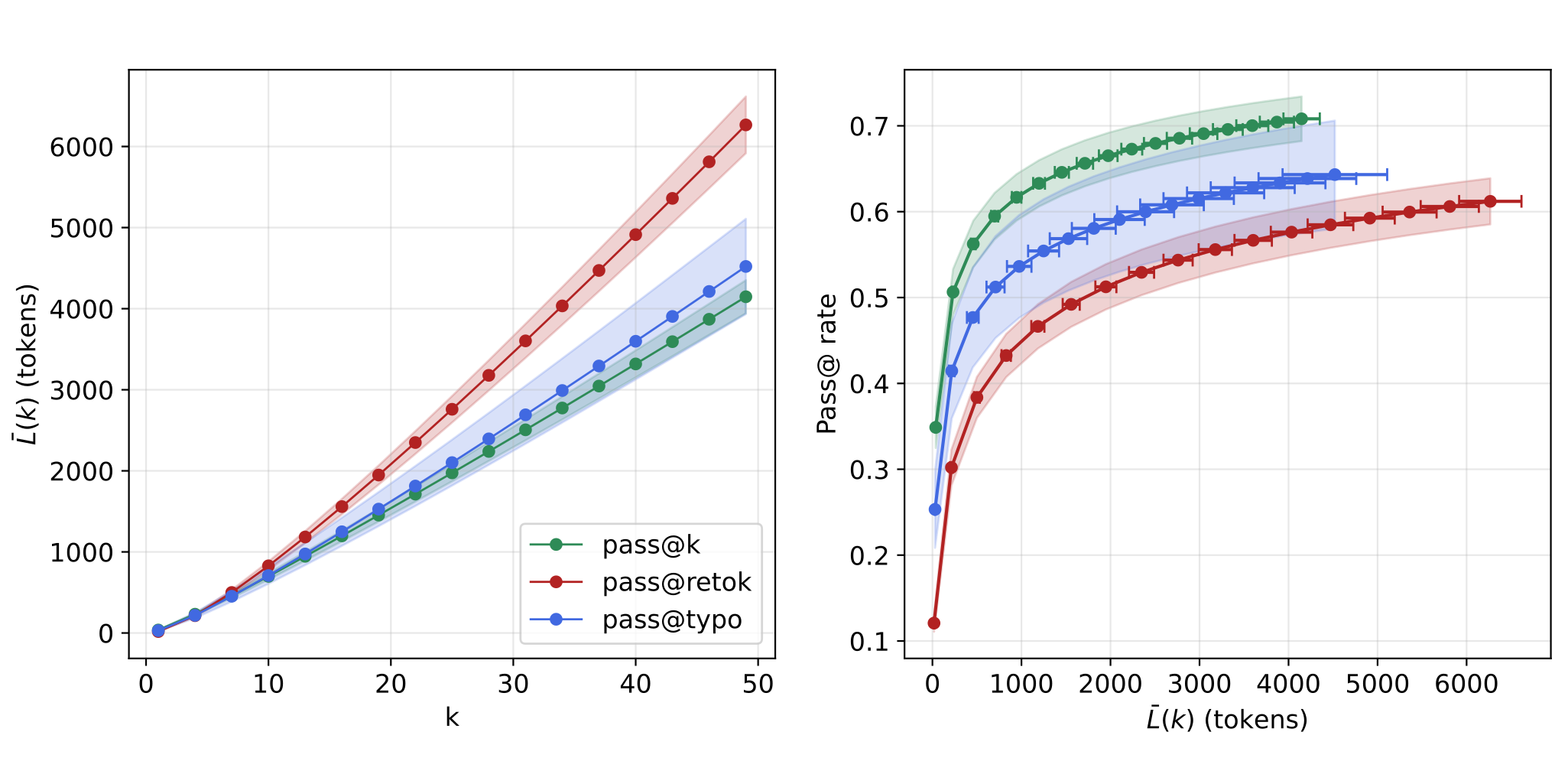}
  \caption{\textbf{Matched-budget comparison on HumanEval.} Left: Expected prompt-token cost $\bar{L}(k)$ as a function of sample count $k$ for temperature sampling, retokenization sampling, and typo sampling. Retokenization is most expensive because alternative segmentations typically lengthen the prompt. Right: pass rate versus expected prompt tokens. Temperature sampling gives the best compute frontier, typo sampling is intermediate, and retokenization is least efficient under this budget, indicating that its main value is complementary diversity rather than raw cost-effectiveness.}
  \label{fig:compute_budget}
\end{figure}

Raw $\passk$ and $\passretok$ curves compare methods at equal numbers of samples, not at equal prompt budgets. This distinction matters because retokenization typically lengthens the prompt, while temperature sampling leaves the prompt fixed and expends compute only through additional decoded outputs. Figure \ref{fig:compute_budget} therefore compares temperature sampling, retokenization, and typo sampling under an explicit prompt-token budget on HumanEval. Here, $\bar{L}(k)$ denotes the expected \textit{prompt} token cost for each sampling process at $k$ samples. The left panel shows the cost side of this comparison, where temperature sampling is the cheapest in prompt tokens, typo sampling is intermediate, and retokenization is the most expensive, likely because equivalent segmentations usually expand the input length. The right panel shows the resulting accuracy frontier under matched prompt budget. Temperature sampling dominates, typo sampling is second, and retokenization is least compute-efficient of the three. We therefore do not claim that retokenization is a drop-in replacement for conventional sampling under strict budget matching. Its value is complementary: it offers a semantics-preserving source of diversity that can recover solutions output-side sampling misses, even if the present compute frontier still favors canonical temperature sampling.

\section{Conclusion}

Equivalent segmentations of the same byte string define a latent degree of freedom that tokenized language models do not treat as irrelevant. Our results show that modern LLMs learn substantial, but incomplete, segmentation invariance: retokenized prompts usually preserve task understanding, yet still perturb the internal computation enough to generate new successful trajectories. This makes retokenization scientifically useful as a probe of compositional understanding and practically interesting as a semantics-preserving source of inference-time diversity.

The picture that emerges is not one of exact symmetry. Behaviorally, this residual sensitivity is most useful on open-ended domains such as code generation, where multiple correct trajectories exist and equivalent segmentations can steer the model toward different algorithmic families. At the same time, matched-budget comparisons make clear that retokenization is not currently a more compute-efficient substitute for conventional temperature sampling. Whether these findings apply to models trained on other languages remains to be explored, and it is unknown whether specific retokenizations help or hurt in specific tasks. Moreover, while we particularly use a stochastic retokenization procedure under a fixed vocabulary, different segmentation policies may yield different tradeoffs between robustness and diversity.

The broader implication is that tokenization should be understood not just as preprocessing, but as part of the model's computational interface to text. As long as deployed LLMs remain tokenized, the space of equivalent segmentations offers a controlled way to study how invariance and sensitivity coexist in sequence models, and how that tradeoff shapes both capability and robustness.

\begin{ack}
We acknowledge helpful conversations with Roberto Avalos, David Berman, Andrey Gromov, Matt Kleban, Noam Levi,  Ilya Nemenman, Sean Ridout and Marat Freytsis. We acknowledge the support and computing resources provided by the HyPER C3 computing cluster at Emory University.

TC acknowledges the support of Emory University.

\end{ack}

\bibliographystyle{plainnat}

\bibliography{arxiv_refs}


\appendix

\section{Experimental Details}
\label{app:exp_details}

This section records the sampling and scoring conventions used to keep canonical, retokenized, and typo-based comparisons aligned across datasets.
\subsection{Stochastic Retokenization}
\label{app:retok}

To generate a non-canonical tokenization $E^{\mu}(s)$ of a string $s$, we first generate the canonical tokenization $E^{0}(s)$ using the model tokenizer. Each resulting canonical token is then independently reconsidered for re-segmentation with a rate $p_{retok}$. Re-segmenting involves finding a random segmentation of a token using shorter tokens that must exist in the tokenizer vocabulary. To do this, we use an existing algorithm used in \citep{zheng_broken_2026} that samples uniformly from all valid ways of segmenting the original token. This is achieved by treating token segmentation as a recursive decomposition problem and sampling paths in the segmentation tree with probabilities  proportional to the number of valid completions, ensuring that each valid segmentation is equally likely. Tokens not selected for re-segmentation ($1 - p_{retok}$) are left unchanged. The final token sequence is formed by concatenating the (possibly re-segmented) tokens in order. By tuning $p_{retok}$, we smoothly interpolate between fully deterministic canonical tokenization ($p_{retok}=0$) and fully stochastic tokenization applied to every token ($p_{retok}=1$).

We apply this procedure only to task content tokens. Special tokens, chat-template markers, and other control symbols remain in canonical form. This ensures that any measured effect is attributable to segmentation of the user-visible task text rather than corruption of the prompt wrapper.

\subsection{Typographical Perturbations}
\label{app:typo}

As an input-side baseline, we also evaluate typo-based perturbations. These runs are constructed analogously to the retokenization runs: for each problem we create multiple perturbed copies of the prompt, vary a corruption probability, and decode from each perturbed prompt using the same model and generation settings as in the corresponding retokenization experiment. The goal is not to study a particular typo taxonomy, but to provide a comparison class of small surface-form perturbations that change the visible text rather than only its segmentation.

For typo-based perturbations, we vary a corruption probability parameter $p_{\mathrm{typo}}$ analogously to the retokenization rate $p_{retok}$. Given a prompt, we first identify the eligible tokens in the perturbable region, namely tokens whose decoded text contains at least one mutable alphabetic character. We then draw a number of token-level typo operations from a binomial distribution with parameter $p_{\mathrm{typo}}$, sample that many eligible character tokens, and apply a random local typo within each selected token by replacing one randomly chosen mutable character by a keyboard-adjacent substitute or a doubled-letter variant (e.g., a $\rightarrow$ q,w,s,z,x,aa; no insertion, deletion, or transposition operators are used). Thus, increasing $p_{\mathrm{typo}}$ increases the expected number of perturbed tokens in the prompt, while keeping the perturbation local and sparse at small $p_{\mathrm{typo}}$. This makes $p_{\mathrm{typo}}$ the typo analogue of $p_{retok}$: both parameters interpolate between the unmodified prompt and progressively more perturbed variants, but in the typo case the perturbation changes the visible surface form rather than only its segmentation.

The scope of the perturbation is dataset dependent. For MMLU and standard GSM8K, typos are injected only into the natural-language problem text, leaving system prompts, answer-format markers, and other control tokens unchanged. For HumanEval and GSM8K Python, we apply typos only within the docstring region of the prompt. We do not perturb the function signature, indentation, or surrounding Python scaffold. This restriction is important: it keeps the task well-formed as a code-completion problem and ensures that any observed changes are attributable to altered natural-language instructions rather than to syntactic corruption of the program context.

\subsection{Calculating pass@ metrics}\label{sec:passatmetrics}

Each question $i$ from a dataset of size $N$ is replicated $r=30$ times for a fixed $p_{retok}$, and stochastic tokenization is applied independently to each replicate by re-segmenting canonical tokens with probability $p_{retok}$. For $p_{retok}$ = 0, this replication is skipped since it generates the canonical tokenization. We sweep over 6 values of $p_{retok}$ $\in  \{0.0,0.2,0.4,0.6,0.8,1.0\}$, obtaining $R = (r \times 5 + 1) $ tokenizations for each prompt. The resulting tokenized inputs are then passed to the model, and inference is performed using greedy decoding (i.e., no output sampling). Thus, for each question $i$, we record how many generations (out of $R$) produce the correct answer. This defines ${\rm pass@retok}$, directly analogous to ${\rm pass@k}$, except that diversity arises from stochastic tokenization rather than output sampling. All experiments are run with the same model, prompt formatting (specific to the benchmark dataset), and decoding parameters, differing only in the retokenization probability. We report ${\rm pass@retok}$ as a function of the number of re-tokenized trials. To calculate ${\rm pass@k}$, we use top-p sampling ($p=0.9$, $temperature = 1$) to generate $R$ generations per question $i$.

For computing ${\rm pass@retok}$, we follow almost identical logic. In this case, $n$ is the number of retokenizations, including the canonical tokenization, and $c$ is the total number of these which, under greedy decoding, generate the correct answer.

We use a dataset specific prompt formatting and answer parsing, as described in sections below.

\subsection{Knowledge Retrieval}

Massive Multitask Language Understanding (MMLU, introduced in \citep{hendrycks_measuring_2021}) is a standard evaluation benchmark comprises of multiple choice questions from a wide variety of subjects. In our experiments we use a subset of 1000 questions sampled randomly from the MMLU dataset. We format each question as a zero-shot multiple-choice prompt, with the model required to output one of the possible answer choices ${A,B,C,D}$. Given an input like this,

\begin{center}
\fbox{
\begin{minipage}{0.90\linewidth}
\ttfamily\small
\textless{}im\_start\textgreater{}system\\
You are a helpful assistant. For the following question, return the answer only, without any additional reasoning or explanation. \textless{}im\_end\textgreater{}

\textless{}im\_start\textgreater{}user\\
Question: $(\mathbb{Z}, *)$ is a group with $a * b = a + b + 1$ for all $a, b \in \mathbb{Z}$. The inverse of $a$ is\\
A.\ 0\\
B.\ -2\\
C.\ $a - 2$\\
D.\ $(2 + a) * -1$\\
Answer:\textless{}im\_end\textgreater{}

\textless{}im\_start\textgreater{}assistant
\end{minipage}
}
\end{center}

we capture the first token generated by the model. If the generated token is not in $\{A,B,C,D\}$, we consider the output to be incorrect, otherwise we compare the generated answer with the true answer option. To calculate ${\rm pass@retok}$, we only re-tokenize the tokens that represent question and its option in the above prompt, leaving the system prompt and special tokens (such as \textless{}im\_start\textgreater{}user) in their canonical form.

Because MMLU is multiple choice with only four valid outputs, naively resampling the answer token can make pass@k-style metrics appear artificially strong even when the underlying reasoning has not improved. We therefore score MMLU from the first answer token only and use the canonical next-token answer probabilities for the temperature baseline discussed above. The appendix random-sampling plot illustrates why this convention is necessary.

\subsection{Mathematical Reasoning}

GSM8K \citep{cobbe_training_2021} is a standard benchmark for evaluating mathematical reasoning on grade-school–level word problems. It consists of free-response questions that require computing a numerical answer rather than selecting from predefined options. In our experiments, we use a randomly sampled subset of 1000 questions from the GSM8K dataset. We do not apply any chat template to the questions as done in the previous datasets.

\begin{center}
\fbox{
\begin{minipage}{0.90\linewidth}
\ttfamily\small
\textless{}pad\textgreater{}\textless{}im\_start\textgreater{}system\\
You are a helpful assistant. \textless{}im\_end\textgreater{}

\textless{}im\_start\textgreater{}user\\
Question:\\
Given a 7-day week, how much does Alex charge for 2 weeks of tutoring if she charges \$12 per day?\\
Answer:\\
\textless{}im\_end\textgreater{}

\textless{}im\_start\textgreater{}assistant
\end{minipage}
}
\end{center}

Given this input, we generate a single completion per prompt and extract the model’s predicted numeric answer from the generated text using a regular expression that selects the final number produced. The prediction is considered correct if this value exactly matches the ground-truth answer.

This exact-match criterion is intentionally strict. It penalizes both arithmetic mistakes and reasoning traces that end with an incorrect final scalar, making GSM8K a useful middle ground between the rigid answer space of MMLU and the open-ended program space of HumanEval.

\subsection{GSM8K Python}

Inspired by previous work \citep{chowdhery_palm_2022}, we also evaluate a code-mediated variant of GSM8K in which each word problem is converted into a Python function-completion task. Rather than asking the model to state the final numeric answer in natural language, we prompt it to write the body of a function whose return value should equal the correct answer. This benchmark sits between GSM8K and HumanEval: the underlying reasoning problem is mathematical, but the output space is programmatic.

Starting from the same GSM8K test questions, we build 1000 problem instances by sampling with a fixed random seed. Each prompt has the form

\begin{center}
\fbox{
\begin{minipage}{0.90\linewidth}
\ttfamily\small
def function():\\
\ \ \ \ """Given a 7-day week, how much does Alex charge for 2 weeks of tutoring if she charges \$12 per day?"""
\end{minipage}
}
\end{center}

where the natural-language question is placed inside the docstring of a Python function. The model must complete the function body. After generation, we extract the code corresponding to \texttt{function} and evaluate it using an automatically constructed test harness. The harness executes the generated function, checks that it returns a finite numeric value, and compares that value against the gold GSM8K answer using \texttt{math.isclose} with absolute tolerance $10^{-6}$. As in HumanEval~\ref{Sec:humaneval}, correctness is therefore functional rather than string-based.

This dataset is useful for our purposes because it separates two roles that are entangled in standard GSM8K. The question itself remains natural language, so retokenization still acts on ordinary text, but the answer must be realized as executable code. This makes GSM8K Python a cleaner bridge between free-response math and full code generation, and it is the benchmark on which our syntactic-diversity analysis can be extended beyond HumanEval. In the typo setting, perturbations are applied only to the docstring region so that the surrounding Python scaffold remains syntactically well-formed.

\subsection{Code Generation}
\label{Sec:humaneval}

HumanEval \citep{chen_evaluating_2021} is a standard benchmark for evaluating code-generation and functional correctness on small programming tasks. It consists of 164 Python problems, each with a function signature, a natural-language docstring, and hidden unit tests. In our experiments, we evaluate on the full set of 164 problems.

\begin{center}
\fbox{
\begin{minipage}{0.95\linewidth}
\ttfamily\small
def generate\_integers(a, b):\\
\ \ \ \ """\\
\ \ \ \ Given two positive integers a and b, return the even digits between a\\
\ \ \ \ and b, in ascending order.\\
\\
\ \ \ \ For example:\\
\ \ \ \ generate\_integers(2, 8) => [2, 4, 6, 8]\\
\ \ \ \ generate\_integers(8, 2) => [2, 4, 6, 8]\\
\ \ \ \ generate\_integers(10, 14) => []\\
\ \ \ \ """\\
\end{minipage}
}
\end{center}

Given each prompt, we generate a single completion and extract the code for the target function. The completion is considered correct if the generated function passes all provided unit tests (i.e., functional correctness at pass@1).

HumanEval is the domain where residual segmentation sensitivity is most behaviorally useful in our experiments. Unlike MMLU, the output space is enormous, and unlike GSM8K, there are many structurally distinct correct solutions. This makes it plausible for semantically equivalent prompt segmentations to steer the model toward different algorithmic families while still preserving task identity.

\section{Benchmark Aggregation and Error Bars}
\label{app:passk_errorbars}

The unbiased estimator for pass-at-k, which we refer to here $\pass(k)$, assumes we uniformly sample from an existing set of $n$ completions, $c$ of which are correct. Then

\begin{align}
{\rm pass}(k) = 1 - \frac{C(n-c,k)}{C(n,k)} , 
\end{align}
where $C(n,k) = \binom{n}{k}$ is the binomial coefficient. For the plotted pass curves, this estimator is applied at the level of individual benchmark tasks rather than after pooling all completions across the benchmark. Concretely, for task $i$ we collect $n_i$ completions and let $c_i$ denote the number of correct ones. The task-level empirical pass rate at budget $k$ is
\begin{align}
{\rm pass}_i(k) = 1 - \frac{C(n_{i} - c_{i}, k)}{C(n_{i},k)}. 
\end{align}
The reported benchmark value is then the mean of these task-level quantities over the $N$ tasks in the benchmark:
\begin{align}
{\rm pass}_{\mathcal{D}}(k) = \frac{1}{N}\sum_{i=1}^{N}\pass_i(k).
\end{align}
Error bars are standard errors over tasks. If
\begin{align}
s_k^2 = \frac{1}{N-1}\sum_{i=1}^{N}\left(\pass_i(k) - \pass_{\mathcal{D}}(k)\right)^2
\end{align}
is the sample variance of the task-level pass estimates at a fixed $k$, then the plotted standard error is
\begin{align}
\mathrm{SE}_k = \frac{s_k}{\sqrt{N}}.
\end{align}
Thus the uncertainty shown in the figures reflects variation across benchmark problems, not variation across repeated generations for a single problem.

The same aggregation logic is used for $\passretok(k)$ and typo-based curves: for each task we first compute the success probability implied by that task's set of variants, and we then average those per-task values across the benchmark. For MMLU temperature curves, where the relevant baseline is the canonical next-token answer probability rather than sampled completions, the task-level quantity is computed from that answer probability and then averaged across tasks in the same way.

\section{Estimating Compute Cost in \method}

The usual $\passk$ estimator treats each draw as equally expensive in prompt tokens, which is reasonable as sampling occurs only from the output distribution. For \method{}, however, the input itself changes length because non-canonical segmentations often use more tokens than the canonical prompt. A compute-aware comparison between sampling methods therefore needs both an accuracy estimator and a prompt-budget model.

To estimate the prompt token cost for task $i$ in a benchmark, let $L_{c,i}$ be the average length of prompts that generate correct completions and $L_{w,i}$ the average length of prompts that generate incorrect completions. For a fixed sample size $k$, the expected total length is
\begin{align}
L_{i}(k) = \sum_{q = 1}^{k} p_{q}^{i} \left( q L_{c,i} + (k - q) L_{w,i}\right)
\end{align}
where $p_q^{i}$ is the probability of drawing exactly $q$ correct completions out of $k$ attempts, given by the hypergeometric distribution 
\begin{align}
    p_{q}^{i} = \frac{\binom{c_{i}}{q}\binom{n-c_{i}}{k-q}}{\binom{n}{k}},
\end{align}
where $c^{i}$ is the number of correct completions among $n$ samples. It is possible to express this in terms of $\passk$:
\begin{align}
    L_{i}(k) = k L_{w,i} \, \pass_{i}(k) + \left( L_{c,i} - L_{w,i}\right) f_{i} k,
\end{align}
where we used the definition $\pass_{i}(k) = \sum_{q = 1}^{k} p_{q}$, and have introduced the total fraction of passing solutions $f_{i} = c_{i}/n$. The benchmark averaged expected length is

\begin{align}
    \bar{L}(k)/k =  \mathbb{E}_{\mathcal{D}}\left[L_{w,i} \pass_{i}(k)\right] + \mathbb{E}_{\mathcal{D}}\left[ \left(L_{c,i} -L_{w,i}\right) f_{i}\right] ,
\end{align}
which assuming independence between the problem dependent variables, gives the approximation
\begin{align}
    \bar{L}(k)/k \approx  \bar{L}_{w}\,  \passk + \left(\bar{L}_{c} -\bar{L}_{w}\right)\bar{f}.
\end{align}
In practice, we estimate prompt-side cost directly from the recovered tokenized prompts for each variant family and then plot pass rate against expected prompt tokens, as in the HumanEval comparison in the main text (Figure~\ref{fig:compute_budget}). This does not capture every source of wall-clock variation, but it is the relevant first-order correction for retokenization because the method changes how much computation is spent on the input context before any completion tokens are generated.

\begin{figure}[H]
  \centering
  \includegraphics[width=0.5
  \textwidth]{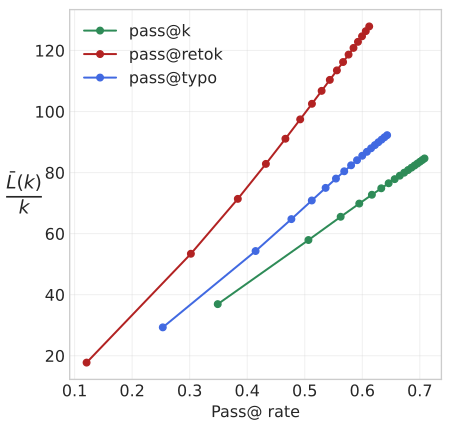}
  \caption{\textbf{Prompt-budget frontier on HumanEval.} The horizontal axis shows benchmark pass rate and the vertical axis shows average prompt-token cost per sample, $\bar{L}(k)/k$, for three sampling procedures: canonical temperature sampling ($\passk$, green), retokenization sampling ($\passretok$, red), and typo sampling ($\passtypos$, blue). Each point corresponds to a different sample budget $k$. Retokenization traces the steepest frontier, indicating that it requires the most prompt tokens per unit gain in pass rate, while canonical temperature sampling remains the most prompt-efficient and typo sampling lies in between.}
  \label{fig:compute_Lk_k}
\end{figure}

\section{Additional Results}

\subsection{Retokenized context helps with next-token prediction}

We probed the emergence of segmentation symmetry in several ways. In this section, we show that a trained model is able to recover useful information in retokenized text, and use it for prediction. In particular, we study whether a model can use a non-canonical segmentation of a passage to improve prediction on the canonical tokenization of the same passage. Following the setup of \cite{sivan_information_2025} for measuring meaningful information rate in communication, we measure the cumulative information content, or surprisal, of a canonically tokenized natural language passage under three conditions: {\bf (1)} no preceding context, {\bf (2)} an identical retokenized version of the passage in context, and {\bf (3)} an identical canonically tokenized version of the passage in context. For conditions {\bf (2)} and {\bf (3)}, we use the following prompt:

\begin{center} 
\fbox{
\begin{minipage}{0.95\linewidth}
\ttfamily\small
The following texts, separated by ----, are identical. {$E'$} ---- {$E^{0}$}
\end{minipage}
} 
\end{center}

Where $E^{0}$ is the canonical tokenization of the passage, and $E'$ is either a retokenization (condition {\bf (2)}) or the canonical tokenization (condition {\bf (3)}). 

\begin{figure}[h]
  \centering
  \includegraphics[width=\textwidth]{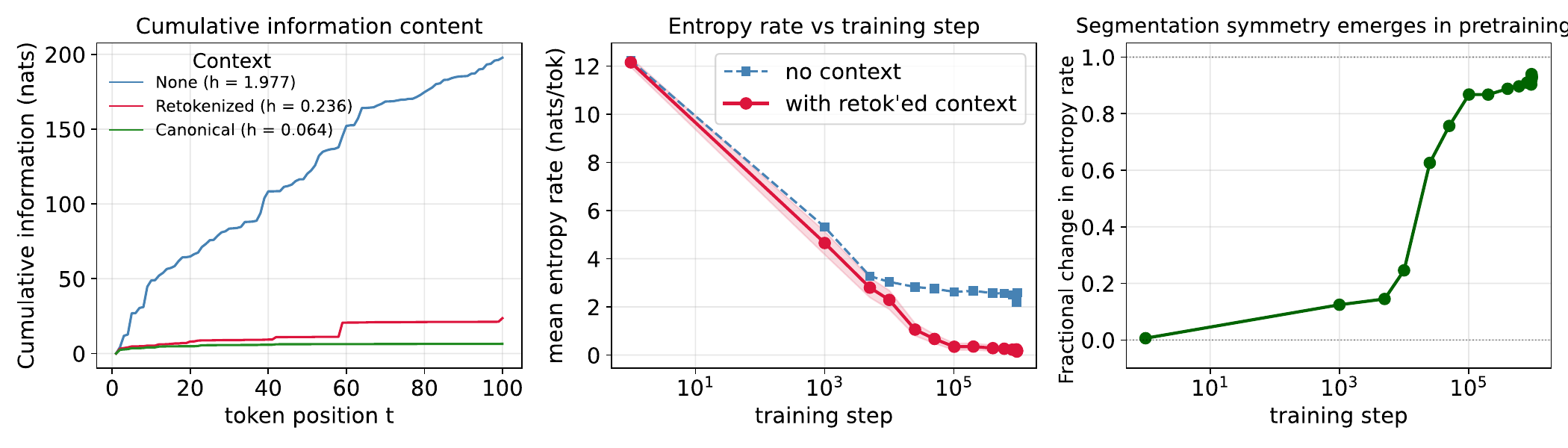}
  \caption{\textbf{Retokenized context sharply reduces surprisal, and this ability emerges during training.} {\bf Left:} Cumulative information content $H_{t}$ \ref{eq:info} of a canonically tokenized passage under three conditions: no context (blue), identical retokenized context (red), and identical canonically tokenized context (green), for Olmo-2-7B Instruct. A retokenized prefix greatly lowers the surprisal of the canonical continuation, approaching the canonical-copy baseline. {\bf Middle:} Across OLMo-2-7B pretraining checkpoints, the mean entropy rate of canonical text remains high irrespective of the context. However, while the no context condition plateaus, the retokenized context continues to decrease. {\bf Right:} The corresponding fractional entropy reduction increases over training, showing the emergence of segmentation-level compositional understanding. Data is computed on 50 passages from the raw variant of Wikitext-103 \cite{merity2017pointer}, using 20 retokenizations per passage, and $p_{retok}=1.0$.}
  \label{fig:fig1}
\end{figure}

To define our measurement mathematically, let $E^{0}(s) = (e_{1}, ..., e_{T})$ be the canonical token sequence for passage $s$, and $W$ be the prefix corresponding to one of the three conditions listed above. The information content of each token in the canonical sequence is given by the logarithm of the probability of that token, conditioned on all past tokens 
\begin{align}
    I_{t}(W) = - \log P\left(e_{t}\mid e_{< t}, W\right).
\end{align}
This is often also called the {\bf surprisal}, since a low value indicates a predictable token, while a high value indicates a low probability and hence ``surprising" token. We measure the cumulative information content, which is the cumulative sum of the information content or surprisal over the token sequence of interest, conditioned on the test condition $W$:
\begin{align}
    H_{t}(W) =  \sum_{s = 1}^{t} I_{s}(W). \label{eq:info}
\end{align}
From this, we get the {\bf entropy rate} $h(W) = H_{T}(W)/T$. The results for OLMo-2-7B are shown in Fig.~\ref{fig:fig1}. If the model understands that the retokenized prefix carries the same content as the canonical sequence, then the surprisal of the canonical continuation should fall sharply relative to the no-context baseline. This is exactly what we observe: after seeing the retokenized text, the information content of the canonical sequence drops substantially, reducing the entropy rate from roughly 2.5 nats/token, typical of natural language, to about 0.06 nats/token. We also track this entropy-rate gap across OLMo-2 pretraining checkpoints \citep{olmo_2_2025}, which lets us follow the emergence of this compositional understanding over training. 

One possibility is that after seeing a retokenized passage, the only entropy that remains is the entropy due to segmentation. If every segmentation was uniformly weighted, the entropy rate would be upper bounded by approximately $\ln2$ ($\approx 0.69$) nats/token, which is well above the observed residual entropy of 0.06 nats/token. However, it is more likely that the distribution over segmentations is nonuniform, and that the distribution is strongly biased toward the canonical segmentation.

\subsection{\texorpdfstring{$P_{fail}$}{Pfail} distribution across datasets}

\begin{figure}[H]
  \centering
  \includegraphics[width=0.8
  \textwidth]{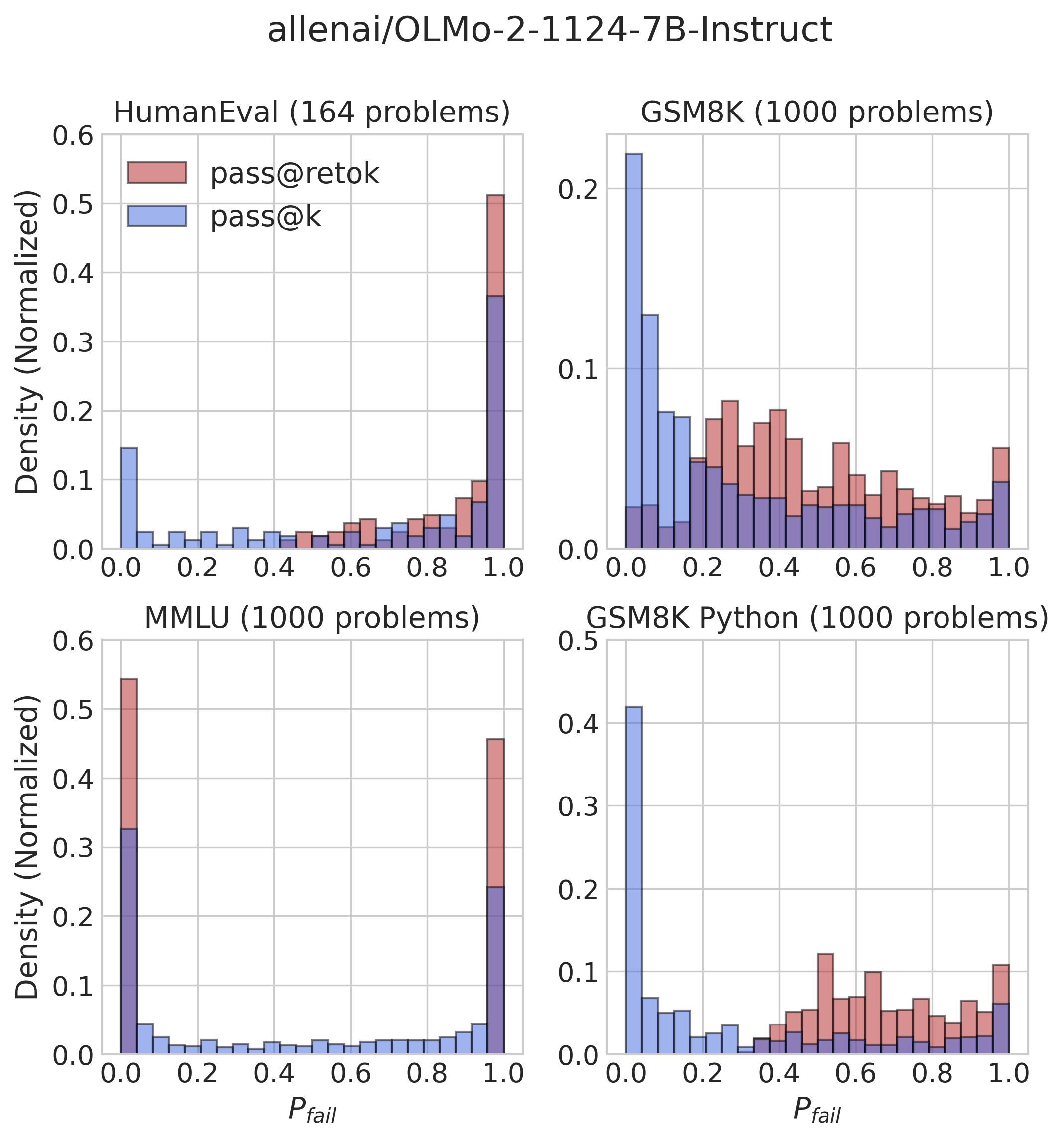}
  \caption{\textbf{Per-problem failure-probability distributions under canonical and retokenized sampling.} For OLMo-2-1124-7B-Instruct, each panel shows the distribution over benchmark problems of $P(\mathrm{failure})$ estimated from canonical output sampling ($\passk$, blue) and retokenization sampling ($\passretok$, red). HumanEval and GSM8K Python show the clearest right-shift under retokenization, indicating that many problems become harder when the same prompt is presented through non-canonical segmentations. GSM8K exhibits a broader spread under retokenization than under canonical sampling, while MMLU remains strongly bimodal because many questions are either almost always solved or almost always failed.}
  \label{fig:pfailures}
\end{figure}

Figure~\ref{fig:fig2} uses aggregate pass curves and $\Delta P$ histograms to show that retokenization affects some problems much more than others. Figure~\ref{fig:pfailures} makes that heterogeneity explicit by plotting the full distribution of empirical per-problem failure probabilities ($P_{fail}$) under canonical output sampling and retokenization sampling. Across datasets, the shape of this distribution reveals whether a benchmark is dominated by easy problems, hard problems, or a broad middle regime. HumanEval and GSM8K Python show the clearest right-shift under retokenization, indicating that many tasks become less reliable when the same prompt is presented through non-canonical segmentations. GSM8K exhibits a broader spread, while MMLU remains strongly bimodal because many questions are either almost always solved or almost always failed. These distributions therefore complement the pass curves by showing not only whether retokenization lowers average performance, but how that loss is distributed across benchmark instances.

\subsection{Retokenization, Typos, and Temperature Sampling}

\begin{figure}[H]
  \centering
  \includegraphics[width=0.95
  \textwidth]{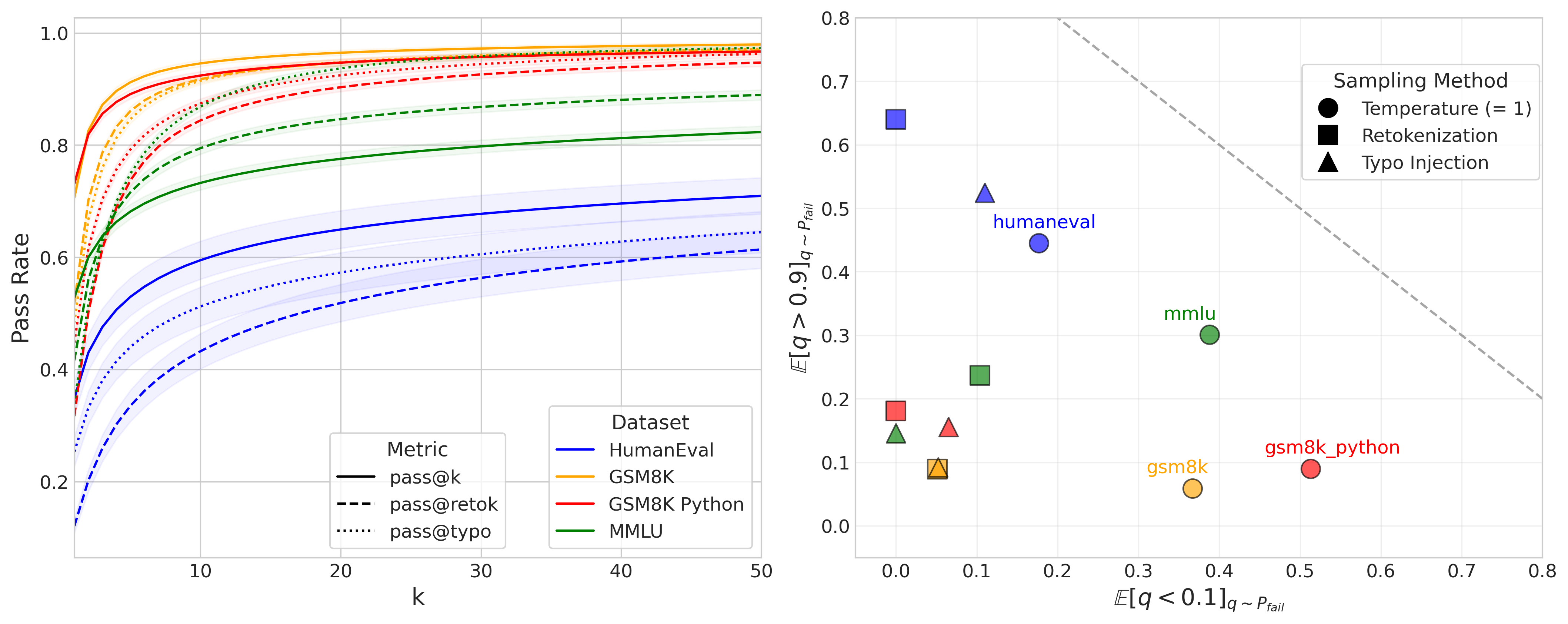}
  \caption{\textbf{All pass curves overlaid across datasets and perturbation families.} Colors denote benchmarks: HumanEval (blue), GSM8K (orange), MMLU (green), and GSM8K Python (red). Line styles denote sampling method: dashed for $\passk$, solid for $\passretok$, and dotted for $\passtypos$. Across HumanEval, GSM8K, and GSM8K Python, conventional output-side sampling is strongest, typo sampling is usually intermediate, and retokenization is weaker but remains competitive. On the right, we plot fraction of hard problems $\mathbb{E}\left[ q > 0.9\right]_{q \sim P_{fail}}$ against fraction of easy problems $\mathbb{E}\left[ q < 0.1\right]_{q \sim P_{fail}}$.}
  \label{fig:fig11}
\end{figure}

Figure~\ref{fig:fig11} overlays the pass curves for canonical output sampling, retokenization sampling, and typo sampling across four benchmarks - HumanEval, GSM8K, GSM8K Python and MMLU. This makes the broad pattern easy to see: on the open-ended tasks, conventional temperature sampling is strongest, typo perturbations usually provide a stronger input-side baseline than retokenization, and retokenization remains the weakest but most controlled search axis. MMLU again stands apart because repeated sampling behaves differently in its small multiple-choice answer space. 

Figure \ref{fig:fig11} provides a compact cross-dataset summary of the ordering between the three sampling procedures. On HumanEval, GSM8K and GSM8K Python, the main pattern is consistent: temperature sampling yields the highest pass curves, typo perturbations provide a stronger input-side baseline than retokenization, and retokenization remains the cleanest but weakest search axis. For MMLU, the limited four-choice answer space makes repeated sampling behavior qualitatively different. This overlay therefore reinforces the main text claim that retokenization is best understood as a controlled and complementary source of diversity rather than a universal replacement for conventional sampling.

\subsection{OLMo-2 Pass Rates on HumanEval at Different Checkpoints and Model Sizes}

\begin{figure}[H]
  \centering
  \includegraphics[width=1.0 \textwidth]{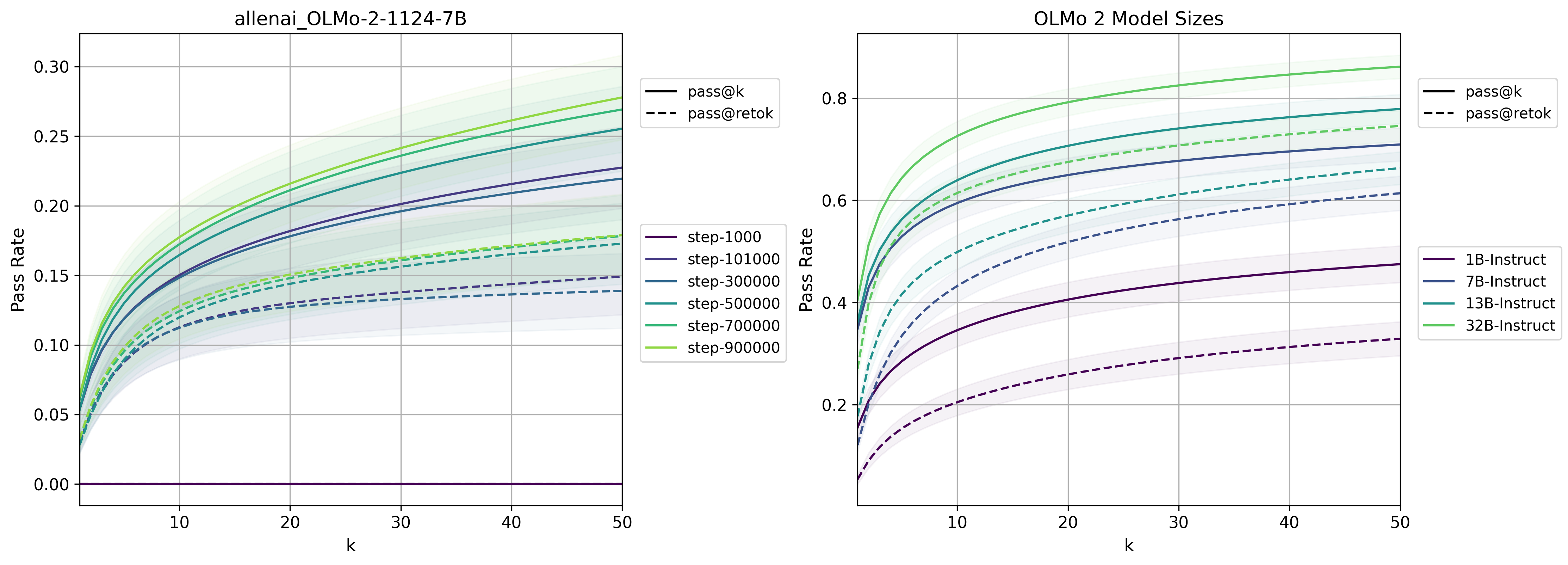}
  \caption{\textbf{Retokenization robustness across OLMo-2 checkpoints and model sizes on HumanEval.} \textbf{Left}: within the OLMo-2-1124-7B pretraining run, later checkpoints improve both canonical $\passk$ (solid) and retokenized $\passretok$ (dashed), indicating that robustness to equivalent segmentations increases with training progress. \textbf{Right}: across OLMo-2 Instruct model sizes, larger models achieve higher pass curves under both sampling procedures, while the gap between canonical and retokenized performance remains model dependent.}
  \label{fig:figpassat_combined}
\end{figure}

\subsection{OLMo-2 Pass Rates Across Retokenization Probabilities \texorpdfstring{$p_{\mathrm{retok}}$}{pretok}}

\begin{figure}[H]
  \centering
  \includegraphics[width=1.0 \textwidth]{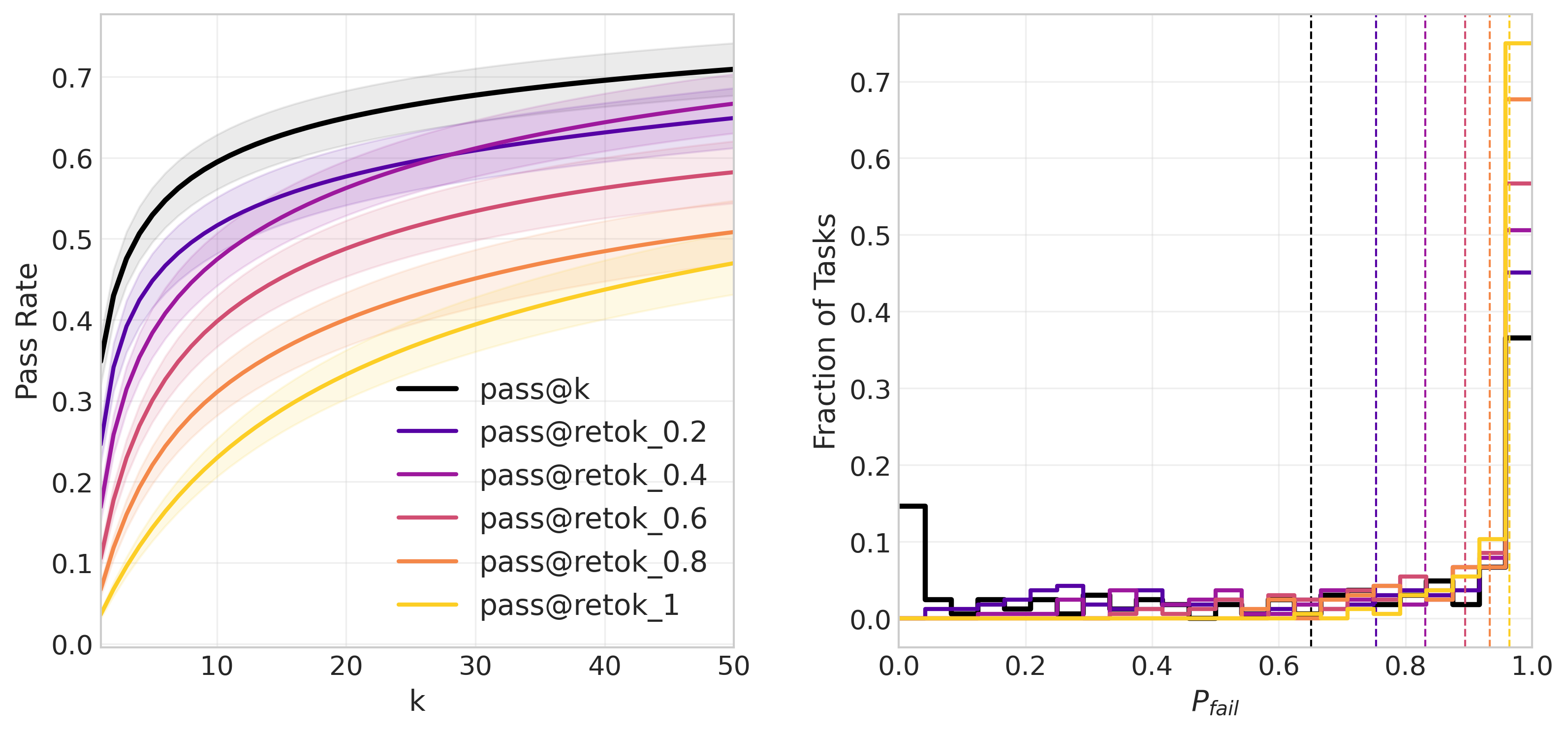}
  \caption{\textbf{HumanEval retokenization curves across perturbation rates.} \textbf{Left}: $\passretok$ curves for HumanEval at different retokenization rates $p_{\mathrm{retok}}$, compared against canonical $\passk$. Moderate retokenization rates preserve more task performance than aggressive retokenization, while still producing diversity. \textbf{Right}: histograms of per-task failure probabilities for the same retokenization rates, showing that increasing $p_{\mathrm{retok}}$ shifts more tasks toward high failure probability. Dashed lines show mean $P_{fail}$.}
  \label{fig:per_p_rates}
\end{figure}

Figure~\ref{fig:per_p_rates} separates the pooled $\passretok$ results by retokenization rate $p_{\mathrm{retok}}$ on HumanEval. This makes the symmetry--sensitivity tradeoff more explicit. Smaller perturbation rates preserve more of the canonical prompt structure and therefore retain higher task performance, while larger rates introduce more aggressive segmentation changes and push more problems toward failure. The left panel shows this as a family of pass curves, where moderate values of $p_{\mathrm{retok}}$ remain competitive over a broad range of $k$ and extreme retokenization is consistently weaker. The right panel shows the same pattern in distributional form: as $p_{\mathrm{retok}}$ increases, the per-task failure-probability histogram shifts toward larger values and the mean failure probability rises. 

\subsection{Segmentation robustness across different model families} 
\label{app:model_families}

\begin{figure}[H]
  \centering
  \includegraphics[width=0.8 \textwidth]{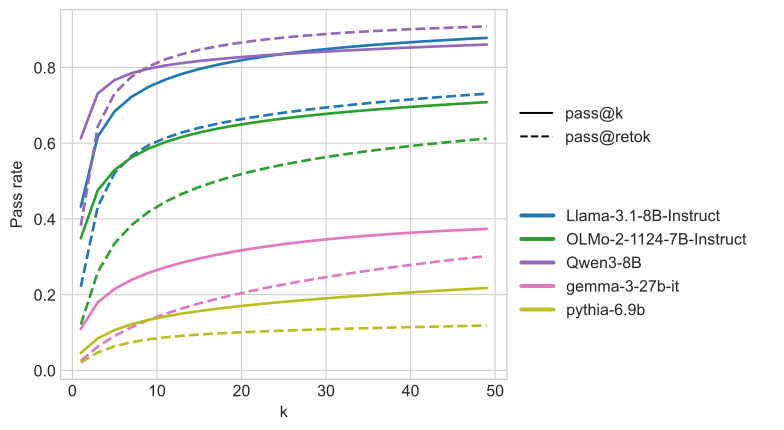}
  \caption{\textbf{Retokenization robustness across model families on HumanEval.} Solid curves denote $\passk$ and dashed curves denote $\passretok$. Stronger models tend to remain stronger under equivalent segmentations, with Qwen3-8B and Llama-3.1-8B-Instruct sustaining the highest retokenized pass curves. }
  \label{fig:fig4}
\end{figure}

Across model families on HumanEval, stronger models also tend to remain stronger under equivalent segmentations (Figure \ref{fig:fig4}). Qwen3-8B \citep{yang_qwen3_2025} and Llama-3.1-8B-Instruct \citep{grattafiori_llama_2024} sustain the highest $\passretok(k)$ curves, while weaker models remain low under both $\passk$ and $\passretok(k)$. At the same time, the gap between canonical sampling and retokenization is clearly model dependent. Qwen3-8B shows the smallest gap, with $\passretok(k)$ remaining very close to canonical $\passk$ across the full range of $k$, whereas Gemma \citep{team_gemma_2025} and Pythia \citep{biderman_pythia_2023} models exhibit substantially larger separations. This suggests that segmentation robustness broadly tracks overall capability, but the extent to which equivalent segmentations preserve useful search behavior is also shaped by model-family-specific inductive biases.

\subsection{Behavioral Plots}
The main text focuses on the aggregate shape of the pass curves. The two figures below isolate two points that are easy to miss in the main presentation: first, how the gap between $\passk$ and $\passretok$ differs by dataset, and second, why MMLU requires special care when interpreted through a pass@k-style lens.

\begin{figure}[h]
  \centering
  \includegraphics[width=0.6 \textwidth]{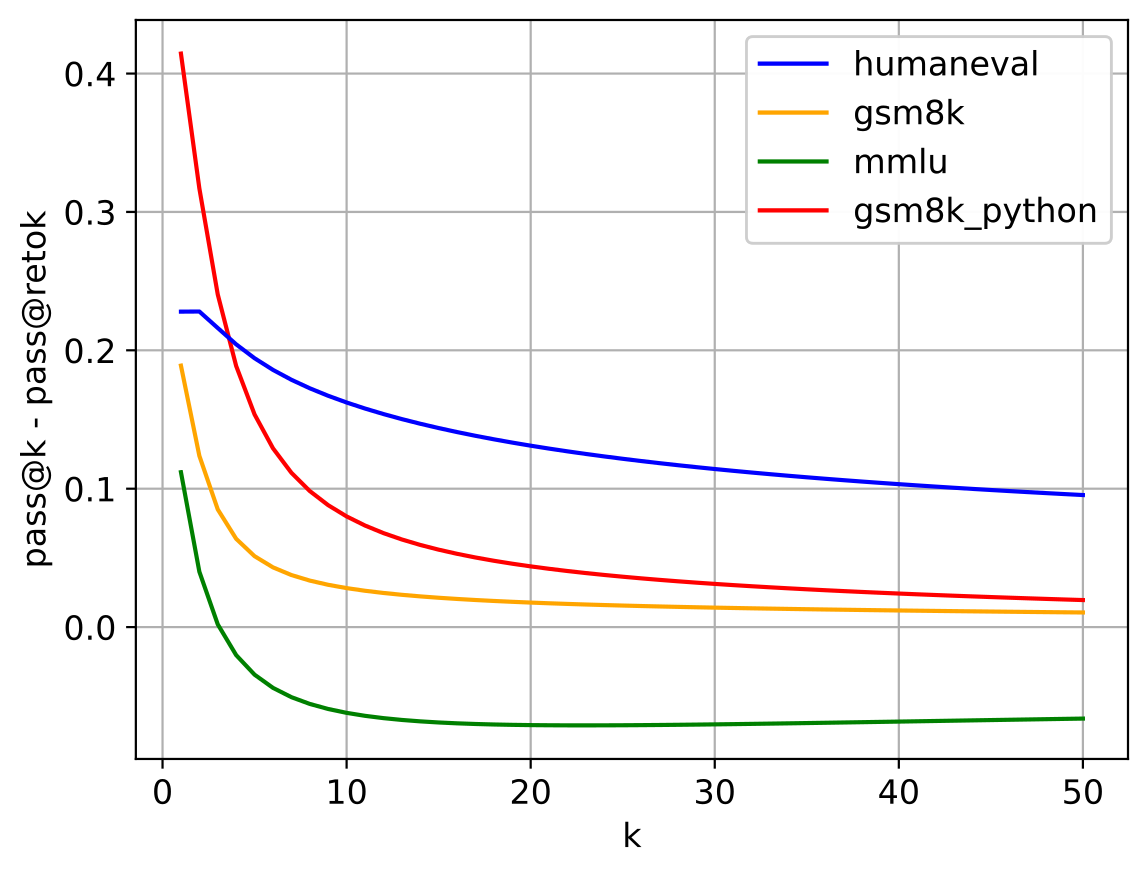}
  \caption{\textbf{Dataset-specific gap between temperature and retokenization sampling.} The plotted quantity is $\passk - \passretok$ as a function of $k$. HumanEval shows the largest persistent advantage for temperature sampling, GSM8K a smaller advantage that gradually shrinks, and MMLU a reversal driven by the peculiar saturation behavior of repeated sampling in a four-choice answer space.}
  \label{fig:fig9}
\end{figure}

Figure \ref{fig:fig9} plots $\passk - \passretok$ as a function of $k$. HumanEval exhibits the largest persistent positive gap, GSM8K a smaller one that steadily shrinks, and MMLU a negative gap. The sign flip on MMLU is not evidence that retokenization is universally stronger there. Instead it reflects the unusual geometry of repeated sampling in a four-choice answer space, where output-side resampling can saturate very quickly.

\begin{figure}[h!]
  \centering
  \includegraphics[width=0.8 \textwidth]{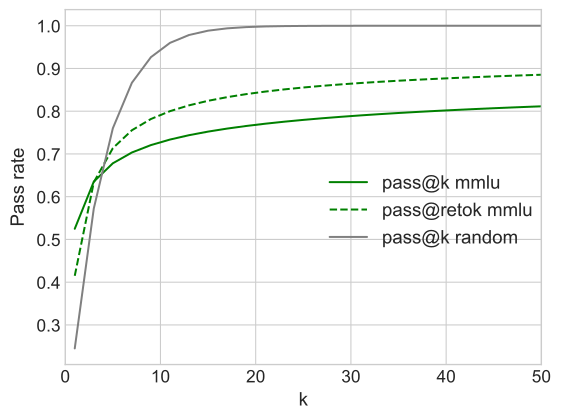}
  \caption{\textbf{Why MMLU needs a special temperature baseline.} The gray dashed curve shows the pass curve obtained by repeatedly sampling answer letters directly from the canonical next-token distribution, while green compares standard $\passk$ and $\passretok$ on the same benchmark. Because the answer space is so small, repeated next-token sampling drives pass rate toward 1 very quickly; this is why MMLU temperature comparisons should be based on the canonical answer probabilities rather than unconstrained multi-token generations.}
  \label{fig:fig10}
\end{figure}

Figure \ref{fig:fig10} makes this saturation issue explicit. If we repeatedly sample answer letters from the model's canonical next-token distribution on MMLU, the induced pass curve approaches 1 extremely quickly. That curve is much stronger than either free-form sampling or retokenization, but it is also much less informative because it mainly reflects repeated draws from a tiny answer set rather than new reasoning trajectories. This is why our MMLU temperature baseline is defined from the canonical answer probabilities of the greedy $p=0$ run rather than from unconstrained multi-token generations.

\subsection{Generating Passing Retokenizations via Iterative BPE Merging}

The subset of initial passing retokenizations on the HumanEval benchmark is a fraction of the total number of initial retokenizations:
a total 20 across all five problems. To increase this subset, we attempt to use these passing retokenizations to generate more.
To accomplish this, we begin with the initial retokenization for a given problem and iteratively apply one BPE merge of
tokens from the tokenizer's merge list (thus ensuring the generated tokens are present in the
vocabulary). We then prompt the model with 
the new retokenization on the same problem and see if the retokenization generates a passing solution using the 
HumanEval unit test suites \cite{chen_evaluating_2021}. If the solution passes the test, we accept the merge, generating a
passing retokenization, and apply the next merge in the merge list. If the solution fails, we
reject the merge, returning to the retokenization before the merge, and attempt to apply the next
merge in the merge list. The process stops once we reach the canonical tokenization or we run out
of valid merges for the retokenization. No problems in our dataset ever reached the canonical
retokenization. For problems with many initial 
retokenizations that generated passing solutions
(HumanEval problems 26 and 159), we perform this process for each initial retokenization
individually. Interestingly, this process worked very well for HumanEval problems 26, 84, and 115, 
generating, collectively, 459 new passing retokenizations, but was not able to generate new passing 
retokenizations for the other problems. The results of this process are shown in {Table~\ref{tab:passing}}.

\begin{table}[h]
\centering
\begin{tabular}{lrrr}
\toprule
\textbf{Problem} & \textbf{Initial passing} & \textbf{Generated passing} & \textbf{Total passing} \\
\midrule
HumanEval/26  & 13 & 374 & 387 \\
HumanEval/72  &  1 &   0 &   1 \\
HumanEval/84  &  1 &  34 &  35 \\
HumanEval/115 &  1 &  51 &  52 \\
HumanEval/159 &  4 &   0 &   4 \\
\bottomrule
\end{tabular}
\caption{Number of passing retokenizations before and after iterative BPE merge generation,
out of 256 initial retokenizations per problem. ``Generated passing'' counts additional
retokenizations produced by the iterative merge process starting from each initial passing
retokenization.}
\label{tab:passing}
\end{table}

\section{Experimental Compute Resources}

Most experiments were run on single H100 and A100 GPUs. The bulk of compute resources were expended in retokenization, temperature and typo sampling procedures. Across all models used in this study, the wallclock runtime for each sampling process was approximately 6 hours for each benchmark dataset (with parameter defined in Appendix~\ref{app:exp_details}). The syntactic diversity calculations were run using 32 Intel Xeon Platinum 8559C cpus for a total wall clock time of 2 hours. Compute cost calculation were run on CPU as well for less than a minute.

\section{Licenses for existing assets}
\label{app:licenses}

We cite the original creators and sources for all models and datasets used in our experiments. The licenses for these artifacts are listed below.

\begin{table}[h]
\centering
\small
\begin{tabular}{lll}
\toprule
\textbf{Artifact} & \textbf{Source / Owner} & \textbf{License} \\
\midrule
OLMo 2 7B & Allen Institute for AI & Apache License 2.0 \\
Qwen3 8B & Qwen Team & Apache License 2.0 \\
Llama 3.1 8B & Meta & Llama 3.1 Community License \\
Pythia 6.9B & EleutherAI & Apache License 2.0 \\
Gemma 3 27B & Google DeepMind & Gemma Terms of Use / Gemma license \\
HumanEval & OpenAI & MIT License \\
MMLU & Hendrycks et al. & MIT License \\
GSM8K & OpenAI & MIT License \\
WikiText-103 test split & Salesforce Research & Creative Commons Attribution-ShareAlike License \\
\bottomrule
\end{tabular}
\caption{Licenses for models and datasets used in our experiments.}
\label{tab:licenses}
\end{table}

\end{document}